\documentclass[10pt,twocolumn,letterpaper]{article}

\usepackage{wacv}
\usepackage{times}
\usepackage{epsfig}
\usepackage{graphicx}
\usepackage{amsmath}
\usepackage{amssymb}
\usepackage{amsfonts} 
\usepackage{nicefrac}       
\usepackage{microtype}      
\usepackage{lipsum}
\usepackage{graphicx}
\usepackage{adjustbox}
\usepackage{subcaption}
\usepackage{multirow}
\usepackage{wrapfig}
\usepackage{xr}

\makeatletter
\newcommand*{\addFileDependency}[1]{
  \typeout{(#1)}
  \@addtofilelist{#1}
  \IfFileExists{#1}{}{\typeout{No file #1.}}
}
\makeatother

\newcommand*{\myexternaldocument}[1]{%
    \externaldocument{#1}%
    \addFileDependency{#1.tex}%
    \addFileDependency{#1.aux}%
}
\myexternaldocument{supplementary}

\usepackage{array}
\newcolumntype{H}{>{\setbox0=\hbox\bgroup}c<{\egroup}@{}}

\def\num#1{\numx#1}\def\numx#1e#2{{#1}\mathrm{e}{#2}}

\def\wacvPaperID{339} 

\wacvfinalcopy 

\ifwacvfinal
\def\assignedStartPage{1} 
\fi

\ifwacvfinal
\usepackage[breaklinks=true,bookmarks=false]{hyperref}
\else
\usepackage[pagebackref=true,breaklinks=true,colorlinks,bookmarks=false]{hyperref}
\fi

\ifwacvfinal
\else
\fi

\begin{document}

\title{Agree to Disagree: When Deep Learning Models With Identical Architectures Produce Distinct Explanations}


\author{Matthew Watson, Bashar Awwad Shiekh Hasan, Noura Al Moubayed\\
Durham University\\
Durham, UK\\
{\tt\small \{matthew.s.watson,bashar.awwad-shiekh-hasan,noura.al-moubayed\}@durham.ac.uk}}

\maketitle

\begin{abstract}
Deep Learning of neural networks has progressively become more prominent in healthcare with models reaching, or even surpassing, expert accuracy levels. However, these success stories are tainted by concerning reports on the lack of model transparency and bias against some medical conditions or patients' sub-groups. Explainable methods are considered the gateway to alleviate many of these concerns. In this study we demonstrate that the generated explanations are volatile to changes in model training that are perpendicular to the classification task and model structure. This raises further questions about trust in deep learning models for healthcare. Mainly, whether the models capture underlying causal links in the data or just rely on spurious correlations that are made visible via explanation methods. We demonstrate that the output of explainability methods on deep neural networks can vary significantly by changes of hyper-parameters, such as the random seed or how the training set is shuffled. We introduce a measure of explanation consistency which we use to highlight the identified problems on the MIMIC-CXR dataset. We find explanations of identical models but with different training setups have a low consistency: $\approx$ 33\% on average. On the contrary, kernel methods are robust against any orthogonal changes, with explanation consistency at 94\%. We conclude that current trends in model explanation are not sufficient to mitigate the risks of deploying models in real life healthcare applications.
\end{abstract}

\section{Introduction}

Deep Learning (DL) applications in healthcare have recently enjoyed a series of successes, with DL models performing on par with human experts leading to the US Food \& Drugs Administration (FDA) to approve 64 DL based medical devices and algorithms as summarised in a recent survey \cite{benjamens2020state}. Whilst these results demonstrate that the trained models are able to perform well on the selected performance criteria, this is not enough for models to reach widespread adoption in practice. This is particularly true in the healthcare domain where it is imperative that the DL models used must be both transparent and explainable, in order to ensure that the relevant stakeholders (patients, medical practitioners) can place their trust in the model, and to help prevent ``catastrophic failures" \cite{danton2020identifying,Hatherley478}.

The ultimate aim of a DL model in highly sensitive applications, such as healthcare, is to capture the underlying causal inter-relationships that medical professionals learn through experience to use for classification. Such a model would be robust to spurious correlations and changes in model training perpendicular to the classification task. Without this level of robustness there will be no trust for its use in the real-world. Current DL training methods often fail to satisfy this requirement, as robustness/trust is yet to be an intricate part of the evaluation and optimisation of said models \cite{damour2020underspecification,nagarajan2020understanding}. An egregious recent example can be seen in certain pneumonia diagnosis models, where it has been shown that the models learned to detect regions (e.g. a metal token placed by the radiologists) of the chest x-ray (CXR) image that indicated which hospital the sample was from, rather than the regions of the image that were causally linked to pneumonia. Despite this, the model still achieved a reasonable ROC-AUC of 0.773 as, incidentally, some hospitals had higher rates of pneumonia than others and so image origin was a good predictor of pneumonia \cite{zech2018variable}. Since the model relied disproportionately on spurious correlations that are not causally linked to pneumonia, it was unable to generalise to unseen data outside of training hospitals.

Recent theoretical and experimental work has demonstrated the challenge of generalisation for DL models and their vulnerability to small changes in the data \cite{dziugaite2020search}. Ensemble models, where multiple, slightly different models work together to make a final prediction, have been proposed to alleviate these issues \cite{hansen1990neural,pang2019improving}. However, while these techniques can improve the robustness of models, they are rarely inherently explainable and do not necessarily capture causal relationships. Additionally, a fundamental requirement of trustworthy models is the interpretability of their decisions. The development of explainable DL techniques to date use either model agnostic post-hoc or model specific approaches. However, the quality of explainable methods is still very difficult to quantify and is designed to be truthful to the model not the data \cite{holzinger2017what,yeh2019infidelity}.

This paper explores the limits of explainable machine learning which highlights fundamental problems in the training and generalisation of neural networks. In particular, we demonstrate how the noise learned by a deep learning model can change significantly when factors such as the random seed, initial weights or even training set order are changed (whilst all other variables remain the same). We propose a measure of the consistency of explanations to quantify the problem and discuss its impact on the interpretation of the explainable output in relation to the input features importance. We show that even the current state-of-the-art ensemble models present with the same issues, and discuss the implications of these findings on the viability of deploying machine learning models in sensitive application domains \cite{alhassan2018stacked,alhassan2018type,esteva2019guide}.

\section{Generalisation and Underspecification}

With the increased use of ML in general and DL in particular, we are becoming increasingly aware of the limitations of DL models. For example, deep neural networks have been shown to be susceptible to imperceptible changes in the the input \cite{szegedy2014intriguing}, or rely on unexpected parts of the input when making their decisions \cite{beery2018recognition}. There is also an increasing number of concerning scenarios wherein a neural network makes biased decisions, such as face detection models reporting high error rates for faces from ethnic minority groups \cite{buolamwini2018gender,Yucer_2020_CVPR_Workshops}. 

There is a growing concern of applications with profound difference between the training dataset and that used in practice, so much so that the differences in the underlying causal structure of the data leads to the poor performance of the trained model \cite{damour2020underspecification}. Even when models are able to generalise well, there is a lack of understanding of why, for example, SOTA vision models converge and generalise even when trained on unstructured noise \cite{zhang2021understanding}. The picture gets even more complex with recent work suggesting neural networks are immune to the bias-variance trade-off with over-parameterised networks demonstrating a striking absence of the classic U-shaped test error curve \cite{neal2018modern,yang2020rethinking}. Additionally, shortcut learning \cite{geirhos2020shortcut}, or decision rules which work well on standard benchmarks but fail to generalise to more complex situations, has recently been shown to be prevalent across many different machine learning domains. Post-hoc explainable methods have gained traction recently to mitigate the issues with model training by opening, albeit rather partially, the black box of a neural network. However, the quality of explanations produced by these methods is difficult to quantify \cite{yeh2019infidelity}. In \cite{dimanov2020you}, the authors demonstrated the susceptibility of explainable methods to the same type of adversarial attacks to that of the original models. We demonstrate here that the generated explanation can be unstable and inconsistent due to variations in model training that are irrelevant to the classification task.

From their inception, ensemble models that incorporate many, diverse sub-models have been proposed to address the problems of robustness and generalisability \cite{sagi2018ensemble,pang2019improving,wenzel2020hyperparameter}. However, as we will demonstrate they also fail to mitigate the low consistency problem of model explanations. We argue that the lack of understanding of exactly how these deep learning models work \cite{weinan2020towards} and generalise is ultimately preventing us from addressing the aforementioned issues. Understanding how the stochastic nature of the training process affects what properties of the data is captured by the model is fundamental. But recent theoretical and experimental studies to understand the generalisation of neural networks concluded the inadequacy of current measures of generalisation \cite{dziugaite2020search,jiang2020fantastic}.

A closer look at explainable outputs of DL models allows us to understand how the randomness introduced during the training significantly affects the explanation of the model's decisions despite consistent accuracy levels. This raises important questions around the robustness of these models. On the contrary kernel methods, namely SVM, are robust against these changes, suggesting that it is the stochastic nature of deep learning model training that may be causing these issues to arise. We argue that these issues significantly impede our ability to confidently suggest DL models for use in healthcare, as they imply that the models might be relying on spurious correlations in the data leading to models producing inconsistent explanations upon retraining. 

\section{Measuring Explanation Consistency}
\label{sec:consistency}

We argue that consistency of the explanations produced by a model regardless of orthogonal changes to hyper parameters is a strong surrogate to model robustness. Fidelity of explanations on the micro level, i.e. input features, is the basis to quantify explanations \cite{yeh2019infidelity,plumb2018model}. Here, we are validating explainability on the macro level, i.e. the robustness of the produced explanation regardless of changes to model training that are orthogonal to the model architecture, data content, and classification task. Intuitively speaking, the consistency of explanations across model variations engender trust in these models as the end user does not expect changes in the explanation due to an incremental model update. Existing similarity metrics of different model outputs(e.g. cosine similarity, root mean squared error) are ill-suited to this task as they are unable to accurately quantify the small (yet important) changes that are particularly of interest here. The separability of a binary classifier, i.e. training accuracy, is an established measure of changes in model output \cite{gan2014filter} which we adapt here to form the basis to measure consistency within the framework defined next. 


\subsection{A Measure of Consistency}
Given a dataset $X=\{x_1,...,x_N\} \subset R^d$, where $d \in \mathbb{N}$ is the dimension of the sample data, we have a classification task $Y(x_i) \in R^n$, where $n$ is the number of classes in a classification setting. We want to evaluate the consistency of explanation method $E$, where $E(Y(x_i)) \in R^d$ assigns a weight to every input feature based on its influence on $Y(x_i)$.

Assume we have $V$ variations of the model $Y$, which we will indicate as $Y^v, v \in \{1,\dots,V\}$, then we define the explanation separability of any two of these variations as:
\begin{equation}
    S_{(a,b)} = \mathbb{E}_i\Big{[}D\Big{(}E(Y^a(x_i)),E(Y^b(x_i))\Big{)}\Big{]}
\label{eq:sep}    
\end{equation}

where $i \in \{1,\dots,N\}$, and $D$ is a similarity measure between the two explanations provided by $E$ of the output of the two models $Y^a$ and $Y^b$, and $\mathbb{E}_i$ is the expected value. The larger $S_{(a,b)}$ is then the more distinct the explanations produced by the same model architecture under the training conditions, $a$ and $b$. Without loss of generality we assume $S_{(a,b)}$ to be normalised in the range $[0,1]$ and we define consistency as:

\begin{equation}
    C = 1- \frac{\sum_{(a,b)}S_{(a,b)}}{\alpha}
\label{eq:c}
\end{equation}

where $\alpha$ is the number of comparisons made between variations of the trained model. The separability metric $S_{(a, b)}$ should be defined such that when the explanations are completely separable (i.e. $S_{(a, b)} = 1$) then the consistency $C = 0$, and vice-versa.

\subsection{Choosing a Suitable Separability Metric}
\label{sec:choosingConsistency}

The definition of $S_{(a,b)}$ should be determined based on the characteristics of $X$, e.g. data dimension and sparsity, and as such it makes sense that different definitions may be appropriate in different scenarios, as long as it is monotonic in the range $[0,1]$. Multiple definitions could be chosen ranging from information-theoretic measures to statistical metrics of similarity (note that similarity metrics can be modified to fit our definition of $S_{(a, b)}$ by ``flipping" their output to ensure that $S_{(a, b)} = 0$ when $a, b$ are identical). Throughout this paper we use the training accuracy of a binary model, $M_{(a,b)}$, trained to classify between $E(Y^a(x_i))$ and $E(Y^b(x_i))$ for $i \in {1,\dots,T}$, where $T$ is the size of the testing set. Eq.\ref{eq:c} can then be re-written as:
\begin{equation}
    C = 1 - \frac{\sum_{(a,b)}2*|M_{(a,b)}-0.5|}{\alpha}
\label{eq:cacc}
\end{equation}
where $|.|$ is the absolute operator. $S_{(a,b)}$ is set to $2*|M_{(a,b)}-0.5|$ to normalise the classification accuracy and make it more meaningful as separability by measuring its distance from theoretical random baseline. An accuracy $M_{(a,b)} = 1$ means the two explanations are completely separable with $S_{(a,b)} = 1$ and $C=0$, and on the other extreme an accuracy $M_{(a,b)} = 0.5$ means that there is perfect agreement between $a$ and $b$ resulting in $S_{(a,b)}=0$ and $C=1$. However, while we have chosen to use the training accuracy of a binary classifier to measure the distance, $D$, between the explainability values, as noted earlier different distance measures could be used and it may be the case that different distance metrics are suited better to different applications and datasets. When choosing a separability metric, it is important to determine whether the chosen distance metric is sensitive enough to detect the small changes in the explanations that we wish to detect. Each possible consistency metric will have various advantages and disadvantages, and it may be that some are better suited to different scenarios; one of the reasons we have chosen to use a binary classifier is its wide range applicability and intuitive interpretation.

Table \ref{tbl:other-measures} contains the values of different divergence measures that we have tested on 4 CNNs (of identical architecture) trained on MNIST with different random seeds. Jensen-Shannon divergence (JSD) is based upon Kullback-Leibler (KL) divergence, and is a method of measuring the similarity between two probability distributions; making it common in machine learning applications, and a prime candidate for use here. JSD is better suited for measuring separability as it is normalised in the range $[0, 1]$. Its main disadvantage is that it measures the divergence between probability distributions, and not samples drawn from a distribution. This requires us to estimate the distribution of the explainability values for the two models under test. This adds an extra layer of complexity to the calculation, and could lead to errors where differences in the techniques and assumptions used to estimate the probability functions. For our experiments reported in Table \ref{tbl:other-measures} we used Kernel Density Estimation (KDE), a method of estimating an unknown probability density function using a kernel function \cite{parzen1962estimation}, which has produced good results, however this would be entirely problem-dependent, whereas the binary classifier method (e.g., Linear Regression(LR)) discussed in the previous section is more generalisable.

\begin{table}[h]
\centering
\begin{adjustbox}{width=\linewidth}
    \begin{tabular}{cc|cccc}
M1 Seed & M2 Seed & JSD & KS & Wilcoxon & LR \\ \hline
1            & 1            & 0             & 0                 &    0  & 0.5                   \\
1            & 12303        & 0.8062        & 0.9744            & $\num{7.877e+09}$    &  0.973          \\
1            & 15135        & 0.8012        & 0.9690            & $\num{1.738e+10}$ & 0.978             \\
1            & 16959        & 0.7346        & 0.8890            & $\num{2.464e+11}$    & 0.975           \\
12303        & 12303        & 0             & 0                 &  0    & 0.5                   \\
12303        & 15135        & 0.8228        & 0.9913            & $\num{4.350e+08}$ & 0.979             \\
12303        & 16959        & 0.7900        & 0.9567            & $\num{3.316e+10}$  & 0.974 \\
15135        & 15135        & 0             & 0                 &   0   & 0.5                   \\
15135        & 16959        & 0.8122        & 0.9810            & $\num{6.611e+09}$ & 0.975             
\end{tabular}
\end{adjustbox}

\caption{Table reporting the Jensen-Shannon divergence, 2 sample Kilmogorov-Smirnov and Wilcoxon signed-rank \textit{test statistics} on the SHAP values from a small subset of the MNIST CNNs tested. The p-values for all hypothesis tests were calculated as 0. Kernel Density Estimation was used before calculating the Jensen-Shannon divergence of the explanations. LR is the accuracy of Logistic Regression classifiers trained on the SHAP values, as used throughout this paper as $M_{(a,b)}$.}
\label{tbl:other-measures}
\end{table}

Statistical hypothesis tests that are designed to test whether two sets of samples are drawn from the same distribution are other candidates. The 2 sample Kilmogorov-Smirnov (KS) test is a two-sided test for the null hypothesis that the 2 sets of samples are drawn from the same continuous distribution \cite{pratt1981concepts}. Using the KS test as a separability measure has the benefit of having a solid statistical underpinning, but we encounter problems when carrying out the test. While we can accurately compute the test statistic (reported for a small set of model in Table \ref{tbl:other-measures}), we cannot compute the associated p-values, preventing us from accurately completing the hypothesis test. In all of our experiments (except those where we were testing a model against itself, where we calculated a test statistic of 0 and p-value of 1), our calculations returned a p-value of 0 (due to technical limitations, we cannot calculate precise enough p-values and so they are rounded down to 0). A similar issue arises when we use the Wilcoxon signed-rank test, which is a non-parametric alternative to the paired t-test which can work on highly non-normal data that works on the null hypothesis that the median differences between pairs of samples are 0. While these results (i.e. calculating a p-value of 0) highlight that our results are highly statistically significant (and hence we can reject the null hypothesis and conclude the explanations are drawn from different distributions), we cannot use results from hypothesis tests to quantify to what degree the explanation's from two models are separable (i.e. we will be unable to infer if one architecture produces more consistent explanations than another), whereas our results with a binary LR classifier allow us to do so. This is not to say that JSD or KS/Wilcoxon hypothesis tests are entirely unsuited to use as a basis for the consistency measure. In this work we have focused our experiments on image data, where input contains a large number of features; applications where fewer features are used might alleviate the technical issues mentioned above. In these cases, it may be appropriate to use one of these measures. However, our choice of a binary classifier is easy to use in any scenario, to any dataset and is easy to interpret and quantify.

\begin{table*}[t]
\centering
\begin{adjustbox}{width=\textwidth}
    \begin{tabular}{c|c|ccc}
    Model Architecture           & Dataset & Shuffle               & Random Seed            & Dropout \\ \hline
    MLP                          & MNIST & $98.195 \pm 0.9550$   & $98.18 \pm 0.94$       & $98.25 \pm 0.8292$  \\
    SVM                          & MNIST & $93.825 \pm 0.7746$   & $94.218 \pm 0.3943$    & n/a     \\
    Small-CNN                    & MNIST & $98.385 \pm 0.0250$   & $98.345 \pm 0.015$     & $98.3267 \pm 0.0330$   \\
    ADP Ensemble                 & MNIST & $98.5 \pm 0.14$       & $99.0875 \pm 0.2573$   & n/a     \\
    CNN                          & MNIST & $97.5 \pm 0.5$        & $99.2170 \pm 0.0443$   & $99.1580 \pm 0.0595$ \\
    GaborNet                     & MNIST & $95.031 \pm 0.2769$ & $95.034 \pm 0.2742$ & $95.054 \pm 0.2934$ \\
    ResNet18                     & MNIST & $99.083 \pm 0.2514$ & $99.471 \pm 0.0438$ & n/a \\
    Densenet-121                 & MIMIC-CXR & $76.005 \pm 0.8363$   & $75.4535 \pm 1.2539$   & n/a     \\
    Densenet-121 Ensemble        & MIMIC-CXR & $81.98 \pm 0.34$      & $80.8533 \pm 0.5311$   & n/a     \\
    Hyperensemble                & MNIST & n/a                   & $99.32 \pm 0.0082$    & n/a    \\
    \end{tabular}
\end{adjustbox}

\caption{Table reporting mean model accuracy ($\pm$ standard deviation) across variations on the base classification task.}
\label{tbl:accuracy}
\end{table*}

\section{Experimental Setup}

We use two publicly available datasets. MNIST is used for efficient baseline tests, and we then extend our experiments to use the MIMIC-CXR-JPG \cite{johnson2019mimicjpg}. We investigate a wide breadth of different model architectures, explanation methods, and training variations\footnote{Code to reproduce our experiments can be found at \href{https://github.com/mattswatson/agree-to-disagree}{https://github.com/mattswatson/agree-to-disagree}}. For both datasets, we use the recommended train/test/val splits. For reproducibility, the specific hyperparameters used for each experiment can be found in the Supplementary Material.

\textbf{MNIST Experiments:} We experimented with the following variations: \textbf{1) MLP} with two hidden layers of sizes 412 and 512 respectively and a dropout layer, \textbf{2) Small-CNN}, a convolutional neural network of 1 convolutional layer with kernel size 3, followed by a max pooling and fully connected layer, \textbf{3) CNN} two convolutional layers with kernel size 3, using max pooling and fully connected layers in between, \textbf{4) GaborNet}, a Small-CNN network with the first convolutional layer restricted to use Gabor filters (the exact parameters of these filters are learned by the network) \cite{bai2019training}, \textbf{5) ResNet18} \cite{he2016deep} with the first convectional layer modified to take 1 channel inputs and the final output layer to have an output size of 10, and \textbf{6) SVM} with RBF kernel. We also train two ensemble models: \textbf{1) ADP ensemble} \cite{pang2019improving} using the default hyperparameters and consisting of 10 ResNet sub-models, and \textbf{2) Hyperensemble} a hyper-batch ensemble \cite{wenzel2020hyperparameter} using the default hyperparameters with 3 sub-models.

\textbf{MIMIC-CXR-JPG Experiments:} The dataset contains 377,110 chest x-rays (CXRs) images from 227,827 studies \cite{johnson2019mimicjpg}. Each study has up to 14 associated labels denoting the disease(s) which are present in the CXR images. For our purposes, we focus only on images with the Edema label; this gives us a subset of 77,483 images of which $47.2\%$ present with the disease (have a positive label) and the remaining $52.8\%$ do not (have a negative label). We use the labels as presented in the MIMIC-CXR-JPG dataset: these have originally been extracted from free-text radiology reports via the CheXpert tool \cite{irvin2019chexpert,johnson2019mimicjpg}. We use the MIMIC-CXR-JPG dataset to demonstrate the issues raised here on a real-life healthcare application. We focus on the Edema label as otherwise we are left with a multi-label classification problem (as one CXR image may show multiple diagnoses), which would make isolating the source of variation very difficult to guarantee. We chose the Edema label specifically as it provides a large number of images whilst also having largely balanced classes. The scope for experimentation with MIMIC-CXR-JPG is necessarily more limited than that with MNIST, as the data requires more complex networks to gain optimal performance. We follow the same process as CheXNet \cite{rajpurkar2017chexnet}, fine tuning a pre-trained Densenet-121 model. We also train a voting ensemble consisting of 3 pre-trained Densenet-121 models trained on subsets of the training dataset.



On both datasets, we train the models repeatedly. For each run we change the hyperparameters that can lead to variations in the randomness used during training without changing the architecture of the model. We change: \textbf{1)} the random seed used during training, \textbf{2)} the dropout rate used in the networks (where applicable), and \textbf{3)} the order of the training data. It is important to note that the train/test/val splits remain the same, rather it is the order in which the training data is passed to the model during training which changes. The accuracy of the models on the base classification task (i.e. MNIST or MIMIC-CXR) are summarised in Table \ref{tbl:accuracy}. To inspect the consistency of decision explanations as a result of changing these hyperparameters, we use two state-of-the-art explainability techniques: SHAP \cite{lundberg2017unified} and Integrated Gradients (IG) \cite{sundararajan2017axiomatic}. These two techniques were chosen as they represent the most commonly used state of the art feature-attribution explanation methods: I) SHAP is a permutation-based model-agnostic approach, so can be applied to the output of any model II) IG is gradient based making it applicable for all neural networks architectures. We calculate the explanation consistency for each explanation technique per model and dataset taking into account every training variation. A Logistic Regression (LR) classifier is used as the binary model to classify between $E(Y^a(x_i))$ and $E(Y^b(x_i))$ as per Eq. \ref{eq:cacc}. This LR model takes the explanation values (i.e. SHAP values, IG values) of the two models as input, and is trained to classify which model the values originated from. The average training accuracy from 10-fold cross validation of the LR model is used. The higher the accuracy of the LR models, the more separable the explainability values are, suggesting that the two models are placing importance on significantly different parts of the input.

To confirm that the underlying problem lies in the models themselves, and not the explainability techniques used, we calculate three different explanation quality metrics that are designed to ensure the explanations produced accurately represent the models: I) (In)fidelity: is the mean squared error between the explanation multiplied by a (meaningful) change in the input and the difference between the model output when given the original and perturbed inputs. II) Sensitivity:  measures the change in explanations when the input is slightly perturbed, calculating this using a Monte Carlo sampling based approximation \cite{yeh2019infidelity}. III) Explanation Accuracy: is the accuracy of a model on the base task (of the same architecture the explanations were produced from) trained on the produced explanations (for example, for MNIST, can a model be trained on the explanations to classify each explanation into one of the 10 digit classes) \cite{molnar2019interpretable}.


\section{Results and Discussion}

\begin{figure*}[t]
    \centering
    \begin{subfigure}{0.3\textwidth}
        \centering
        \includegraphics[width=\linewidth]{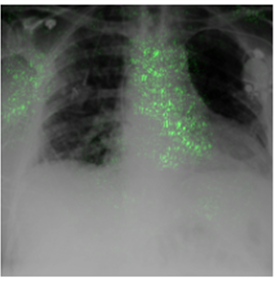}
    \end{subfigure}
    \begin{subfigure}{0.3\textwidth}
        \centering
        \includegraphics[width=\linewidth]{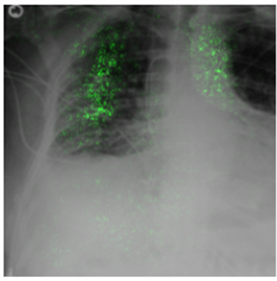}
    \end{subfigure}
    \begin{subfigure}{0.3\textwidth}
        \centering
        \includegraphics[width=\linewidth]{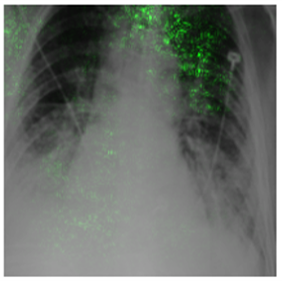}
    \end{subfigure}
    \caption{3 random samples from the MIMIC-CXR-JPG dataset overlayed (in green) with the difference between the normalised SHAP values from two Densenet121 training variations.}
    \label{fig:cxr-shap-diff}
\end{figure*}

\begin{figure*}[t]
     \centering
     \includegraphics[width=\textwidth]{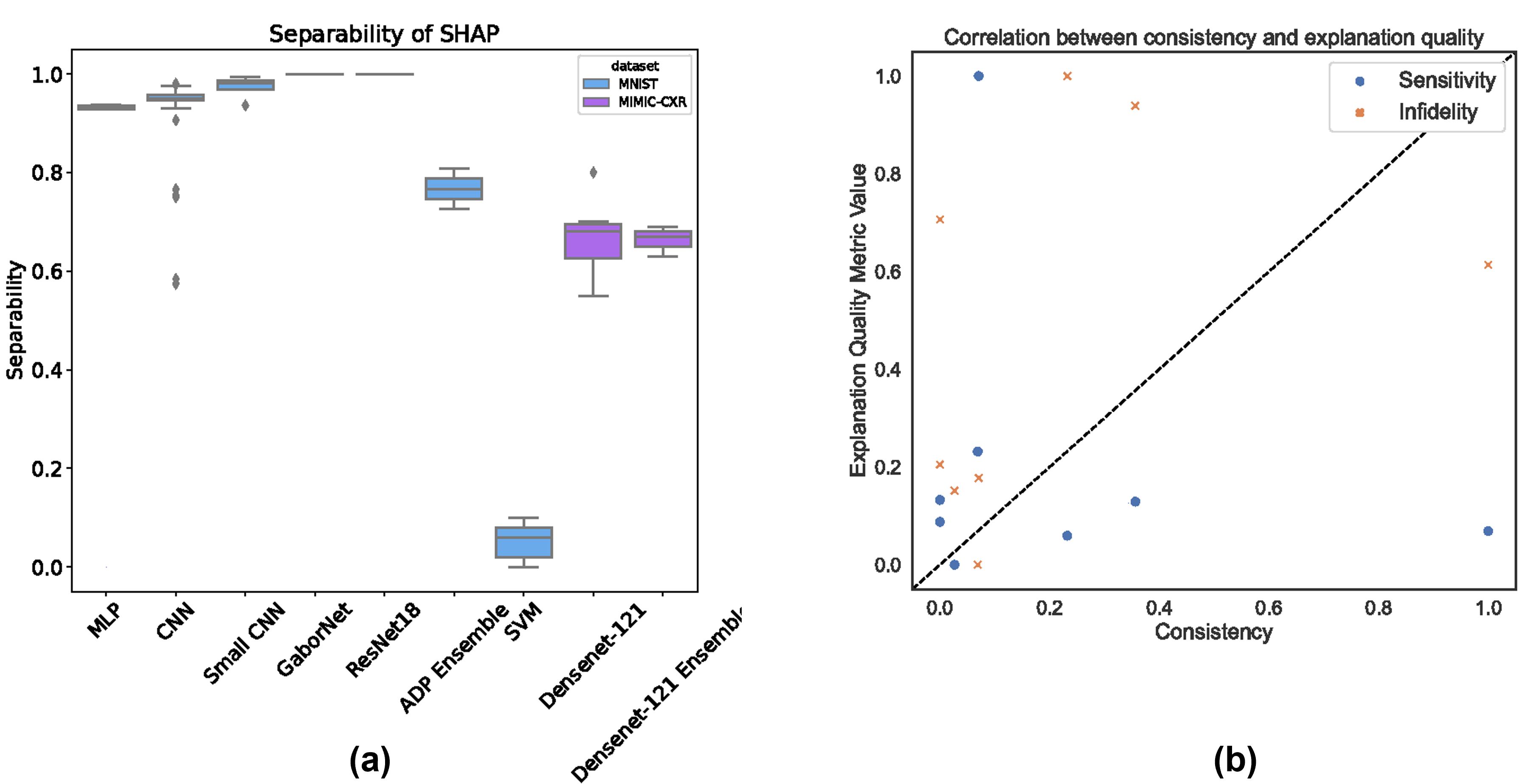}
     \caption{(a) Box plot of $S_{(a,b)}$ for SHAP across all training variations $(a, b)$, for all model architectures tested. (b) Plot of SHAP explanation consistency of model architectures vs. SHAP infidelity and sensitivity of the same models across both MNIST and MIMIC data.}
     \label{fig:boxplots}
\end{figure*}

\begin{figure}[h!]
    \centering
    \includegraphics[width=\linewidth]{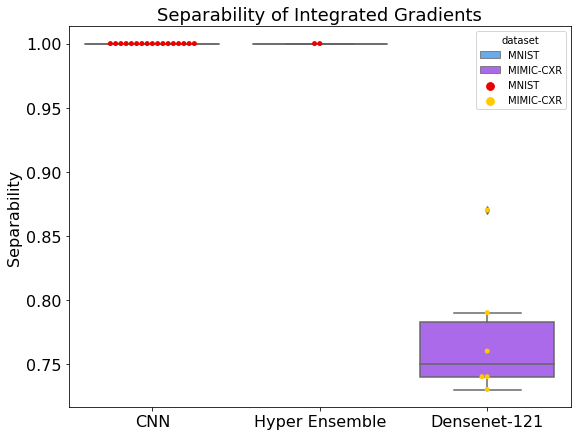}
    \caption{Boxplot of the separability $S_{(a,b)}$ of the Integrated Gradients explanations.}
    \label{fig:boxplot-ig}
\end{figure}

\begin{figure*}[t]
    \centering
    \includegraphics[width=\textwidth]{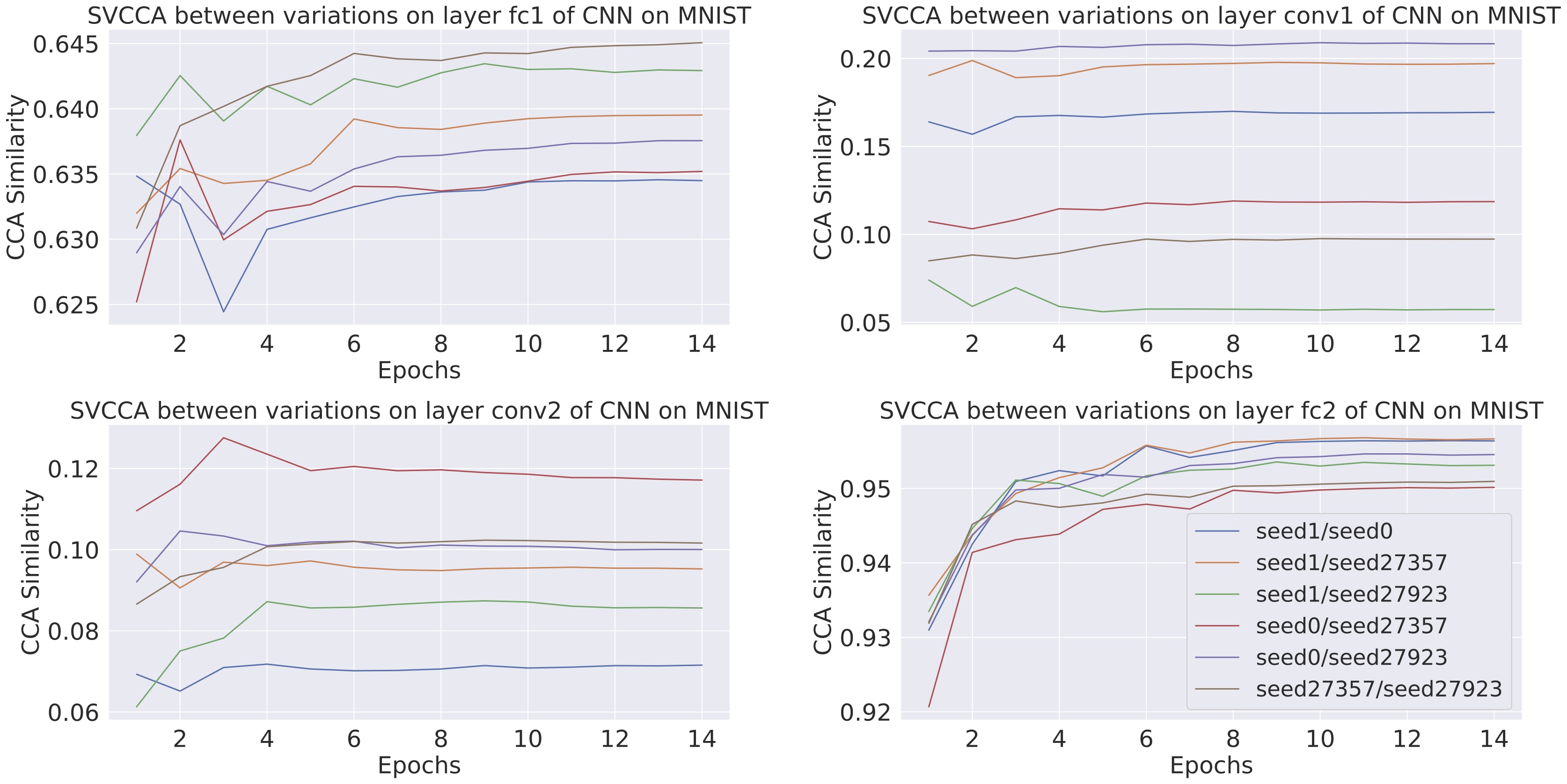}
    \caption{Figures showing the CCA similarity as training progresses between layer parameters. Each coloured line is a separate training variation pair of a CNN trained on MNIST.}
    \label{fig:svcca}
\end{figure*}

\begin{table*}[t]
\centering
\begin{adjustbox}{width=\textwidth}
    \begin{tabular}{c|c|ccccc|c}
    &                            &  \multicolumn{5}{c|}{Consistency} &  \\
    Model Architecture           & Dataset & $\alpha$  & Overall & Shuffle & Random Seed & Dropout & Accuracy \\ \hline
    MLP                          & MNIST & 6 & 0.0668      & 0.062   & 0.066       & 0.0687  & $98.125 \pm 0.9270$        \\
    SVM                          & MNIST & 10 & 0.9444      & 0.96    & 0.94        & n/a     & $94.0556 \pm 0.6213$         \\
    Small-CNN                    & MNIST & 6 & 0.0252      & 0.018   & 0.06        & 0.034   & $98.3486 \pm 0.0360$         \\
    GaborNet                     & MNIST & 12 & 0 & 0 & 0 & 0 & $95.038 \pm 0.2824$ \\
    ResNet18                     & MNIST & 10 & 0 & 0 & 0 & n/a & $99.425 \pm 0.0626$ \\
    ADP Ensemble                 & MNIST & 6 & 0.2193      & 0.192   & 0.233       & n/a     & $99.083 \pm 0.2514$         \\
    CNN                          & MNIST & 12 & 0.0652      & 0.052   & 0.0564      & 0.0914  & $98.9976 \pm 0.5756$         \\
    Densenet-121                 & MIMIC-CXR & 6 & 0.3329      & n/a     & 0.3329      & n/a     & $75.6723 \pm 1.1379$         \\
    Densenet-121 Ensemble        & MIMIC-CXR & 4 & 0.3367      & n/a     & 0.3667      & n/a     & $80.8 \pm 0.7483$            \\ \hline
    CNN  (IG)                    & MNIST & 12 & 0           & 0       & 0           & 0       & $98.9976 \pm 0.5756$         \\
    Hyperensemble (IG)           & MNIST & 2 & 0           & n/a     & 0           & n/a     & $99.32 \pm 0.0082$     \\
    Densenet-121 (IG)            &  MIMIC-CXR & 6 & 0.168       & 0.115     & 0.2033    & n/a     & $75.6723 \pm 1.1379$       \\
    \end{tabular}
\end{adjustbox}
\caption{Table reporting the consistency between training variations for the models tested and the average accuracy of the model architecture on the base classification task. The Shuffle, Random Seed and Dropout columns report the consistency of models when \textit{only} the respective hyperparameter was changed. The Overall column reports the overall consistency of that architecture, taking an average of the consistency across all hyperparameters. $\alpha$ refers to the number of models tested for the overall architecture consistency (see Eq. \ref{eq:c}). Please refer to Table 1 in the Supplementary Material for $\alpha$ values for the shuffle, seed and dropout consistencies.}
\label{tbl:consistency}
\end{table*}

Through visualisation of the explanation differences, we are able to discern whether the lack of consistency between variations is a cause for concern when deploying deep learning models to real-world scenarios. Figure \ref{fig:cxr-shap-diff} demonstrates the change in explanations between two variations of the same Densenet-121 model using SHAP. We see two main sets of differences in the images: \textbf{1)} areas of the image that are clinically significant (e.g. the lungs and the heart), and \textbf{2)} areas in background portions of the image. Those differences that are in clinically relevant to diagnosis can result in significantly reduced trust in the model, as we ideally want a model which has learnt the entire set of causal links present in the data (whereas these differences show that the two models have learnt to look at different sets of causal features). The remaining differences are in the background noise of the images, which suggests that the models are potentially picking up spurious correlations, with each model learning different sets of spurious correlations. Neither of these scenarios is desirable. Examples on Small-CNN trained on MNIST are shown in Figure 1 in the Supplementary Material - similarly to the CXR samples, we can see that the changes in the SHAP values are mainly centered around the areas of the image that are critical for number classification. These results are significant - it suggests both that variations in the training setup of a model changes the importance of the fundamental features that we would expect to be causally linked to the final classification, and on more complex tasks are also changing the spurious correlations learned by models.

Following, we report the accuracy of all models tested on MNIST and MIMIC-CXR-JPG, and the consistency of the explainability methods per model/dataset. Table \ref{tbl:consistency} contains each model architecture's consistency, and a further breakdown of the consistency for the different types of training variation tested. For all model architectures, the degree of consistency is similar regardless of which hyperparameters is changed; this suggests that deep learning models are sensitive to all training hyperparameters, and not just a select few. Figures \ref{fig:boxplots}(a) and  \ref{fig:boxplot-ig} further demonstrate the variation in the separability measure ($S_{(a,b)}$) used to calculate consistency across all models/datasets. These figures show that there is very little consistency of either SHAP and IG for any training variation when used with deep learning models. By contrast, we find that SVMs do not suffer from the same issue as deep learning models, achieving very high levels of consistency across both random seed and training shuffle variations. This provides evidence for our hypothesis that it is the stochastic nature of deep learning model training that may be causing these issues to arise. Figure \ref{fig:boxplot-ig}  shows the boxplot for IG, with even more pronounced separability, which can likely be attributed to how IG is calculated based on the weights of the network. Figure \ref{fig:boxplots}(a) does not show any real link between the size/depth of a network architecture and the separability/consistency. 

Interestingly, both GaborNet and ResNet18 are highly inconsistent. The purpose of Gabor filters in CNNs is to more accurately simulate our biological understanding of human vision, with these filters picking up low-level features. Our results show that the features picked up are inconsistent - intuitively this makes sense, with lower-level features being more prone to smaller changes in the model. The purpose of testing the ResNet18 architecture was to investigate whether overparameterised networks also suffer from this inconsistency problem; as can be seen in Figure \ref{fig:boxplots}(a) and Table \ref{tbl:consistency}, they do. This implies that even models which have many more times the number of parameters than data points are converging to slightly different points on the loss landscape when small hyperparameter changes are made. It also suggests that even high capacity networks, which we would expect to be able to learn the entire set of meaningful features, are in fact either not able to do so. We hypothesise that, although ResNet18 is most likely learning (to some degree) all of the features present in the model, it is applying different weights to the noise present in the model when training hyperparameters are changed. This is not surprising as the overparameterised model has more chance of picking up spurious correlations. Figure \ref{fig:boxplots}(b) shows the correlation, or lack thereof, between explanation consistency and (in)fidelity and sensitivity as measures of the explanation's quality across all experimental settings. Both measures show weak Pearson correlation (0.4 for (in)fidelity and -0.3 for sensitivity). This is not surprising as those metrics are designed to be faithful to the model, not to the underlying data. This disparity between explanation consistency and quality highlights the problem with the use of explanation methods as a surrogate to model transparency. A lower consistency model is less robust and can lead to misinterpretation of model output, hence damaging the confidence of using the model in sensitive domains. To further measure the quality of the SHAP and IG explanation, we also calculate the explanation accuracy for each model, i.e. the accuracy of the model trained on the explanation output of the sample data: a higher accuracy suggests more representative explanations. We report each model's individual explanation infidelity, sensitivity max and accuracy in Table 1 in the Supplementary Material. The weak correlation between the quality metrics and consistency lead to two conclusions: \textbf{1)}  explanation quality metrics are unable to detect inconsistency in the models, and \textbf{2)} if the explanations are indeed faithful to the model, then the only remaining source of inconsistency is the trained model itself, or more precisely the training approach of these models. Ensemble approaches seem to have higher consistency but it is still significantly lower than that of SVMs. We use SVCCA \cite{raghu2017svcca} to inspect the similarity of layer parameters between two training variations, and how these change as training progresses. SVCCA views neurons as their activation vectors, and uses an amalgamation of Singular Value Decomposition and Canonical Correlation Analysis to analyse these representations: we encourage the interested reader to peruse \cite{raghu2017svcca} for a more thorough explanation. Figure \ref{fig:svcca} shows the SVCCA similarity between layers of CNNs trained on MNIST with different random seeds. It shows a high degree of similarity for the final layer, whereas the middle layer (\texttt{conv2}) shows a significant difference. This corroborates our explainability consistency results; the final layers (\texttt{fc2}) are similar and so the models will produce similar outputs, resulting in similar performance levels. Conversely, all other layers are significantly different and so the explanations, which take into account the whole model, are different. In addition, the two convolutional layers show an extremely low degree of similarity between the two models, hence the feature maps learned by these two models are likely also not similar resulting in lower consistency.

\section{Conclusion}

In this paper we introduced a consistency measure of explainable machine learning and demonstrated that deep learning models converge to learn different features when the same model is trained with different random seeds, training set orders and dropout rates. By validating the quality of the explanation techniques used, and using both gradient-based and perturbation-based techniques, we have shown that this is a fundamental problem with deep learning models rather than an issue with the explanations. Additionally, we verified that SVMs are immune to this problem. We argue that there is still significant work that need to be done to build robust trustworthy deep learning solutions in real-life healthcare applications.

\section*{Acknowledgements}
This work is supported by grant 25R17P01847 from the European Regional Development Fund and Cievert Ltd.

\clearpage

{\small
\bibliographystyle{ieee_fullname}
\bibliography{main}
}

\end{document}


\title{Agree to Disagree: Supplementary Material}


\author{Matthew Watson, Bashar Awwad Shiekh Hasan, Noura Al Moubayed\\
Durham University\\
Durham, UK\\
{\tt\small \{matthew.s.watson,bashar.awwad-shiekh-hasan,noura.al-moubayed\}@durham.ac.uk}}

\maketitle

\begin{minipage}{\textwidth}
    \centering\noindent
    \begin{subfigure}{0.19\textwidth}
        \centering
        \includegraphics[width=\linewidth]{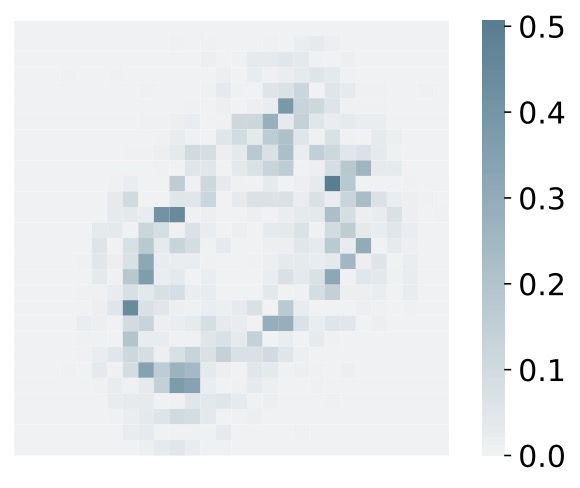}
    \end{subfigure}
    \begin{subfigure}{0.19\textwidth}
        \centering
        \includegraphics[width=\linewidth]{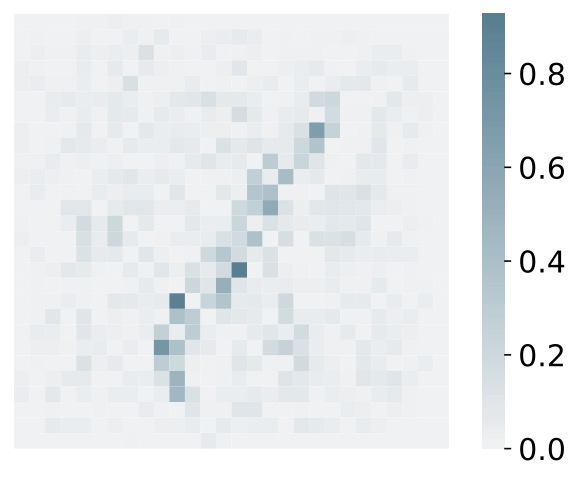}
    \end{subfigure}
    \begin{subfigure}{0.19\textwidth}
        \centering
        \includegraphics[width=\linewidth]{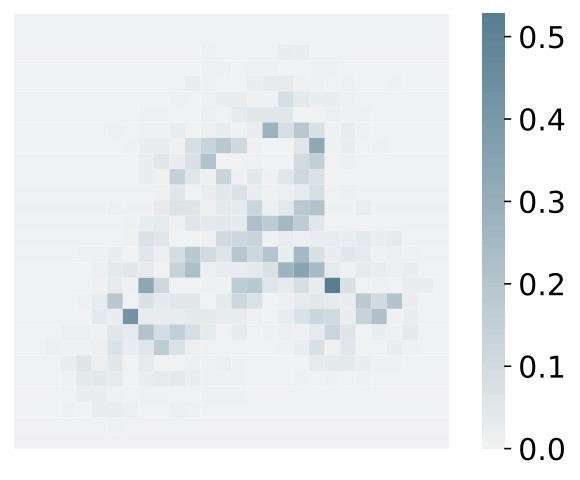}
    \end{subfigure}
    \begin{subfigure}{0.19\textwidth}
        \centering
        \includegraphics[width=\linewidth]{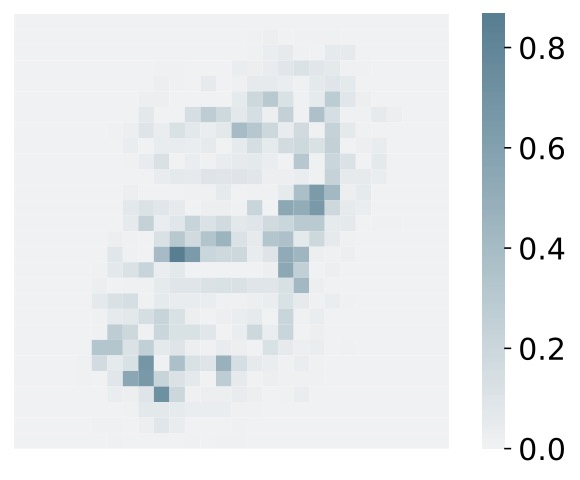}
    \end{subfigure}
    \begin{subfigure}{0.19\textwidth}
        \centering
        \includegraphics[width=\linewidth]{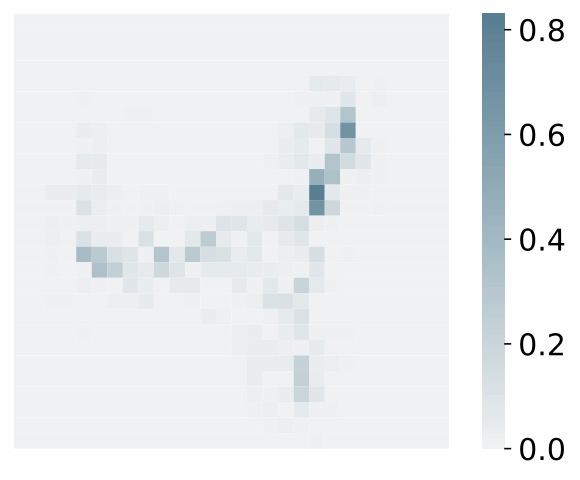}
    \end{subfigure}
    
    \begin{subfigure}{0.19\textwidth}
        \centering
        \includegraphics[width=\linewidth]{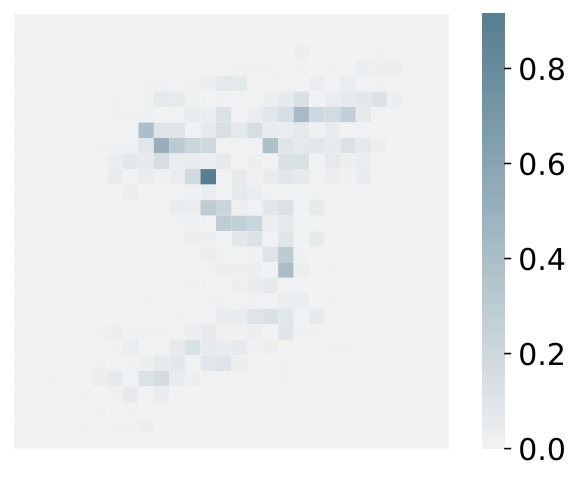}
    \end{subfigure}
    \begin{subfigure}{0.19\textwidth}
        \centering
        \includegraphics[width=\linewidth]{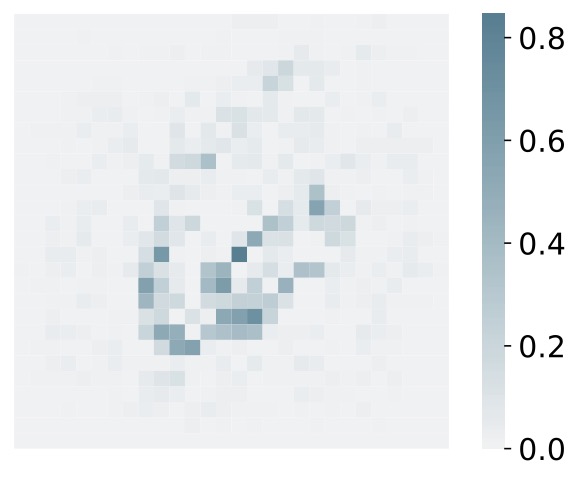}
    \end{subfigure}
    \begin{subfigure}{0.19\textwidth}
        \centering
        \includegraphics[width=\linewidth]{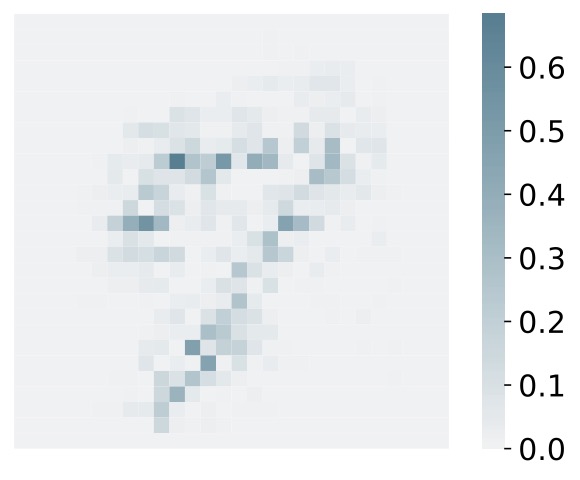}
    \end{subfigure}
    \begin{subfigure}{0.19\textwidth}
        \centering
        \includegraphics[width=\linewidth]{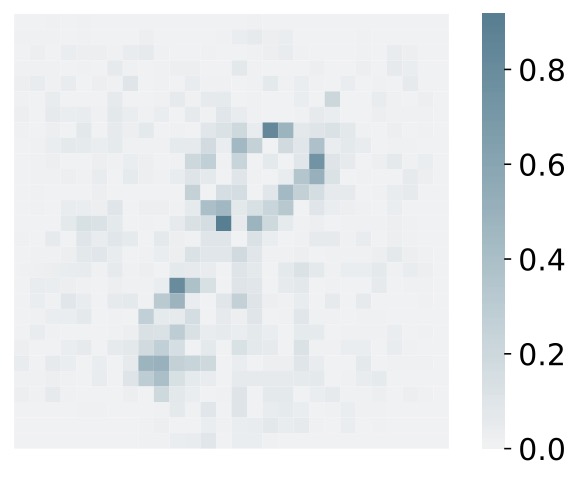}
    \end{subfigure}
    \begin{subfigure}{0.19\textwidth}
        \centering
        \includegraphics[width=\linewidth]{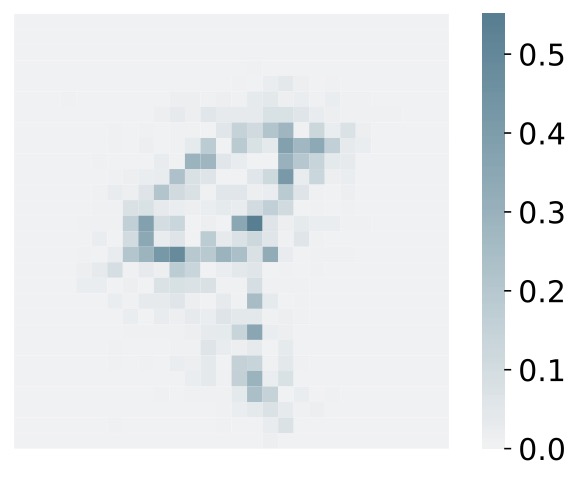}
    \end{subfigure}
    \captionof{figure}{The difference between normalised SHAP values from two CNNs (each trained with different random seeds) for a randomly chosen sample from each MNIST class.}
    \label{fig:mnist-shap-diff}
\end{minipage}

\begin{figure*}[b]
    \centering
    \begin{subfigure}{0.19\textwidth}
        \centering
        \includegraphics[width=\linewidth]{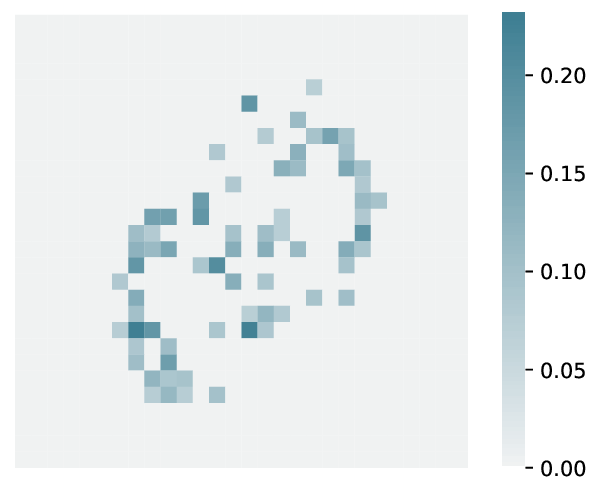}
    \end{subfigure}
    \begin{subfigure}{0.19\textwidth}
        \centering
        \includegraphics[width=\linewidth]{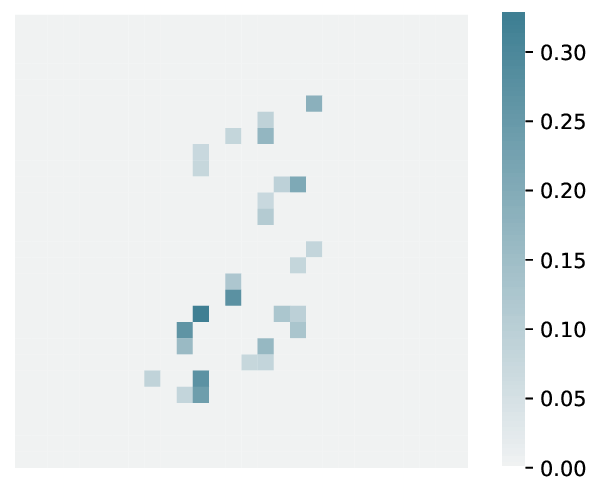}
    \end{subfigure}
    \begin{subfigure}{0.19\textwidth}
        \centering
        \includegraphics[width=\linewidth]{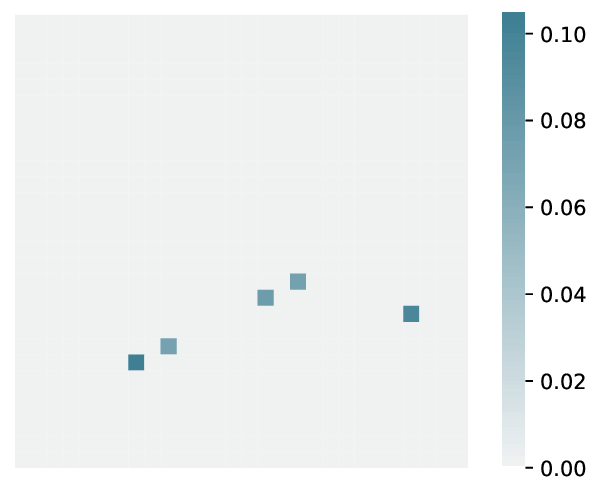}
    \end{subfigure}
    \begin{subfigure}{0.19\textwidth}
        \centering
        \includegraphics[width=\linewidth]{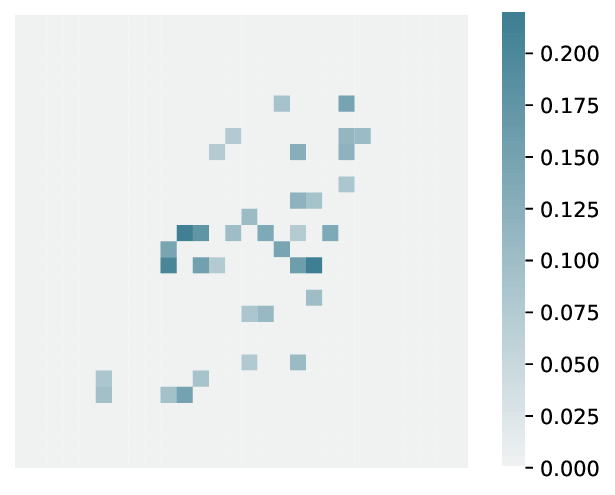}
    \end{subfigure}
    \begin{subfigure}{0.19\textwidth}
        \centering
        \includegraphics[width=\linewidth]{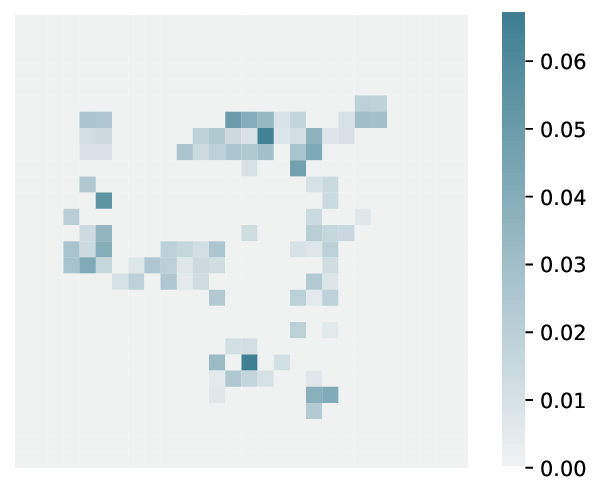}
    \end{subfigure}
    
    \begin{subfigure}{0.19\textwidth}
        \centering
        \includegraphics[width=\linewidth]{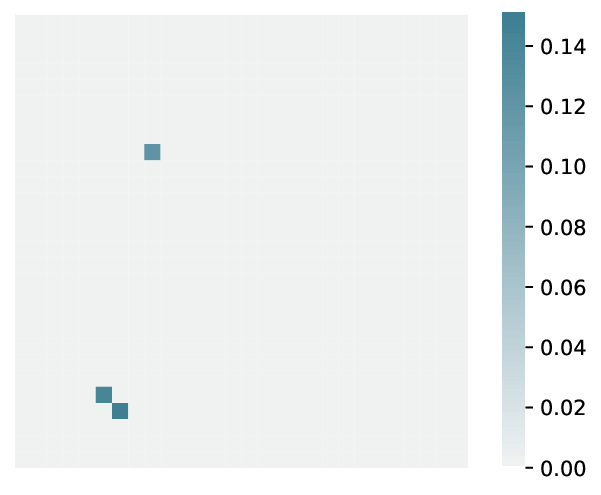}
    \end{subfigure}
    \begin{subfigure}{0.19\textwidth}
        \centering
        \includegraphics[width=\linewidth]{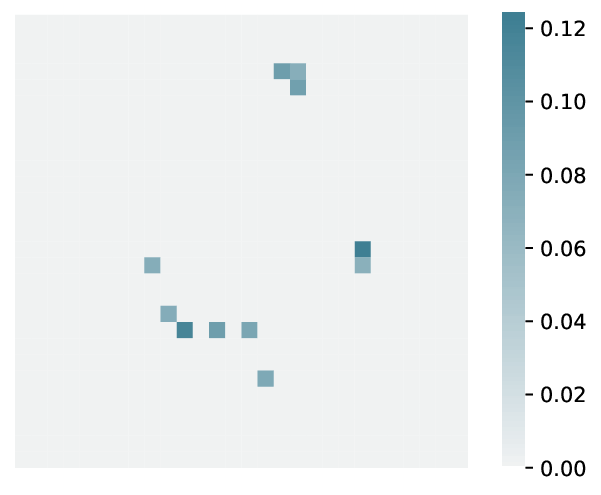}
    \end{subfigure}
    \begin{subfigure}{0.19\textwidth}
        \centering
        \includegraphics[width=\linewidth]{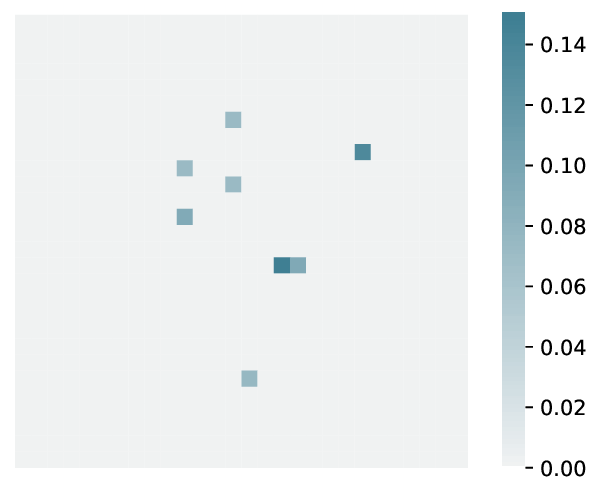}
    \end{subfigure}
    \begin{subfigure}{0.19\textwidth}
        \centering
        \includegraphics[width=\linewidth]{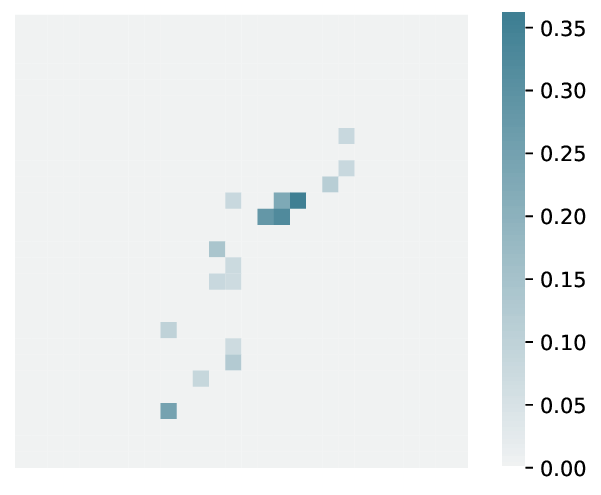}
    \end{subfigure}
    \begin{subfigure}{0.19\textwidth}
        \centering
        \includegraphics[width=\linewidth]{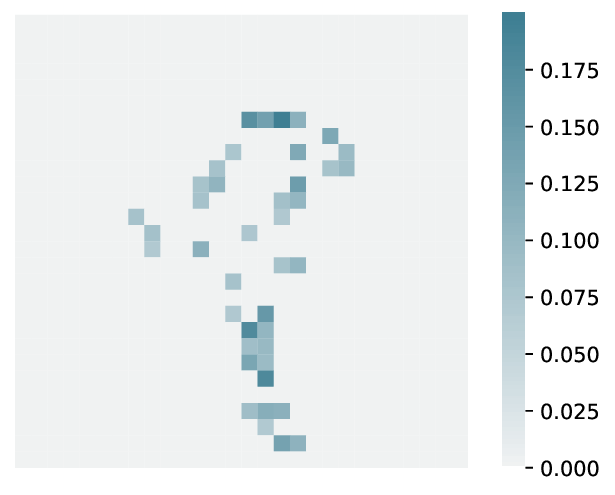}
    \end{subfigure}
    \caption{The difference between normalised SHAP values from two SVMs (each trained with different random seeds) for a randomly chosen sample from each MNIST class.}
    \label{fig:mnist-shap-diff}
\end{figure*}

\begin{table*}[h]
\centering

\resizebox*{!}{0.92\textheight}{
\begin{tabular}{c|c|ccc|ccc}
Model Type                & Dataset & Dropout & Seed  & Shuffle & Explanation Accuracy & Sensitivity & Infidelity \\ \hline
\multirow{12}{*}{\rotatebox[origin=c]{45}{CNN}}& \multirow{12}{*}{\rotatebox[origin=c]{45}{MNIST}} & 0.0     & 1     & False   & 96       & 2.40.       & 0.0019                \\
                                               & & 0.1     & 1     & False   & 97       & 2.04        & 0.0018                    \\
                                               & & 0.2     & 1     & False   & 97       & 2.02        & 0.0020                 \\
                                               & & 0.3     & 1     & False   & 98       & 1.69        & 0.0019                    \\
                                               & & 0.4     & 1     & False   & 98       & 1.53        & 0.0016                    \\
                                               & & 0.25    & 1     & False   & 98       & 1.84        & 0.0016                    \\
                                               & & 0.25    & 12303 & False   & 98       & 1.70        & 0.0014                     \\
                                               & & 0.25    & 15135 & False   & 98       & 1.58        & 0.0020                    \\
                                               & & 0.25    & 16959 & False   & 97       & 1.67        & 0.0018                    \\
                                               & & 0.25    & 20878 & False   & 98       & 1.61        & 0.0020                    \\
& & 0.25 & 79266 & True & 99 & 1.51 & 0.0014 \\
& & 0.25 & 79870 & True & 99 & 1.67 & 0.0011 \\ \hline
                                               
\multirow{6}{*}{\rotatebox[origin=c]{45}{Small-CNN}} & \multirow{6}{*}{\rotatebox[origin=c]{45}{MNIST}} & 0.0   & 1    & False   & 99    &  1.07       & 0.1810                       \\
                         & & 0.2     & 1     & False   & 98                             &  1.00       & 0.1429                       \\
                          & & 0.25    & 1     & False   & 98                            &  1.00       & 0.1521        \\
                          & & 0.25 & 26417 & False & 99                                 &  1.02       & 0.1011 \\
& & 0.25 & 91110 & True & 99 & 1.01 & 0.1174 \\
& & 0.25 & 98281 & True & 99 & 1.01 & 0.1402 \\ \hline
                          
\multirow{12}{*}{\rotatebox[origin=c]{45}{GaborNet}} & \multirow{12}{*}{\rotatebox[origin=c]{45}{MNIST}} & 0.0 & 0 & False & 99 & 1.38 & 0.2808\\ 
& & 0.1 & 0 & False & 99 & 1.41 & 0.2256\\   
& & 0.2 & 0 & False & 99 & 1.42 & 0.1900\\   
& & 0.3 & 0 & False & 99 & 1.44 & 0.1702\\   
& & 0.4 & 0 & False & 99 & 1.46 & 0.1523\\
& & 0.25 & 257 & False & 99 & 1.17 & 0.1489\\  
& & 0.25 & 6339 & False & 99 & 1.34 & 0.2508 \\   
& & 0.25 & 29062 & False & 99 & 1.40 & 0.1683\\   
& & 0.25 & 51303 & False & 98 & 1.45 & 0.2352\\   
& & 0.25 & 17939 & True & 98 & 1.34 & 0.1567\\   
& & 0.25 & 23682 & True & 98 & 1.28 & 0.1190\\  
& & 0.25 & 27442 & True & 99 & 1.31 & 0.1274\\  
& & 0.25 & 53307 & True & 99 & 1.27 & 0.1089\\  \hline

\multirow{10}{*}{\rotatebox[origin=c]{45}{ResNet18}} & \multirow{10}{*}{\rotatebox[origin=c]{45}{MNIST}} & 0.25 & 21609 & False & 99 & 1.15 & 0.7214\\ 
& & 0.25 & 23474 & False & 99 & 0.96 & 0.4426\\
& & 0.25 & 29246 & False & 99 & 2.34 & 0.5284\\
& & 0.25 & 48769 & False & 98 & 0.83 & 0.5007\\
& & 0.25 & 58626 & False & 99 & 1.21 & 0.7121\\
& & 0.25 & 72 & True & 98 & 1.21 & 0.5572\\
& & 0.25 & 1507 & True & 98 & 1.42 & 0.8697\\
& & 0.25 & 4439 & True & 99 & 0.97 & 0.5402\\
& & 0.25 & 10250 & True & 99 & 2.10 & 0.8867\\
& & 0.25 & 21033 & True & 99 & 1.01 & 0.9018\\ \hline
                          
\multirow{6}{*}{\rotatebox[origin=c]{45}{MLP}} & \multirow{6}{*}{\rotatebox[origin=c]{45}{MNIST}} & 0.0 & 1 & False & 99                 &  3.49       & 0.1748 \\
                                               & & 0.2 & 1 & False & 99                 &  5.56       & 0.1573 \\
                                               & & 0.25 & 1 & False & 99                &  4.85       & 0.1508 \\
                                               & & 0.25 & 27833 & False & 99            &  3.76       & 0.1926 \\
& & 0.25 & 72 & True & 99 & 3.39 & 0.1427 \\
& & 0.25 & 79870 & True & 99 & 3.74 & 0.1399 \\ \hline                                        
                                               
\multirow{6}{*}{\rotatebox[origin=c]{45}{Densenet121}} & \multirow{6}{*}{\rotatebox[origin=c]{45}{MIMIC}} & n/a & 2 & False & 99 & 1.5966 & 0.9994 \\ 
                                                      &  & n/a & 3 & False & 99 & 1.5031 & 1.0719 \\
                                                      & & n/a & 4 & False & 99 & 1.5987 & 1.0020 \\    
                                                      & & n/a & 5 & False & 99 & 1.1431 & 0.4659 \\
& & 0.25 & 6 & True & 99 & 1.5122 & 0.9994 \\
& & 0.25 & 7 & True & 99 & 1.6078 & 1.1217 \\ \hline                                                      
                                                      
\multirow{5}{*}{\rotatebox[origin=c]{45}{ADP}} & \multirow{5}{*}{\rotatebox[origin=c]{45}{MNIST}} & n/a & 0 & False & 99 & 1.2187 & 0.9110 \\ 
                                                      &  & n/a & 42 & False & 99 & 1.4250 & 1.4376 \\
                                                      & & n/a & 100 & False & 98 & 1.2297 & 0.9730 \\    
& & 0.25 & 1 & True & 99 & 1.3491 & 0.9912 \\
& & 0.25 & 10 & True & 98 & 1.3100 & 1.266 \\ \hline

\multirow{4}{*}{\rotatebox[origin=c]{45}{DNE}} & \multirow{4}{*}{\rotatebox[origin=c]{45}{MIMIC}} & n/a & 1 & False & 80 & 1.5499 & 1.0357 \\
                                                      &  & n/a & 42 & False & 84 & 1.3709 & 0.7340 \\
& & 0.25 & 4242 & True & 81 & 1.5683 & 0.6510 \\
& & 0.25 & 1000 & True & 82 & 1.6932 & 0.8493 \\ \hline                                                      
\multirow{5}{*}{\rotatebox[origin=c]{45}{SVM}} & \multirow{5}{*}{\rotatebox[origin=c]{45}{MNIST}} & n/a & 30828 & False & 99 & 1.5763 & 0.2070 \\ 
                                                      &  & n/a & 31599 & False & 99 & 1.1686 & 0.9074 \\
                                                      & & n/a & 8253 & False & 99 & 1.0238 & 0.6214 \\
& & 0.25 & 91244 & True & 99 & 1.5439 & 0.5006 \\
& & 0.25 & 79870 & True & 99 & 1.5894 & 0.4823 \\                                                      
\end{tabular}}

\caption{Table reporting explanation quality metrics on SHAP across all model architectures and training variations tested. DNE denotes Densenet-121 Ensemble. Where Shuffle is \texttt{True}, Seed refers to the seed used for shuffling the dataset and not the training seed.}
\label{tbl:qual}
\end{table*}